\documentclass{article}
\usepackage{ijcai17}
\usepackage[utf8]{inputenc}
\usepackage[usenames, dvipsnames]{color}
\definecolor{lred}{RGB}{255,143,128}
\definecolor{env}{RGB}{239,141,34}
\definecolor{agent}{RGB}{12,124,186}
\definecolor{reward}{RGB}{201,45,57}
\definecolor{mygray}{gray}{0.7}
\definecolor{mygray2}{gray}{0.5}

\usepackage{times}

\usepackage[british]{babel}

\usepackage{subcaption}

\usepackage{rotating}

\usepackage{amsmath,amssymb}
\usepackage{fdsymbol}
\usepackage[ruled]{algorithm2e}
\usepackage{graphicx}

\newcommand{\past}{\overleftarrow}
\newcommand{\future}{\overrightarrow}
\newcommand{\bs}[1]{\boldsymbol{#1}}
\newcommand{\mc}[1]{\mathcal{#1}}
\newcommand{\mb}[1]{\mathbb{#1}}
\newcommand{\bmc}[1]{\bs{\mc{#1}}}
\newcommand{\sime}{\sim_\epsilon}
\newcommand{\larr}{\leftarrow}

\DeclareMathOperator*{\argmin}{arg\,min}

\newtheorem{mydef}{Definition}
\newtheorem{myobj}{Objective}
\newtheorem{myprop}{Property}

\title{General AI Challenge \\Round One: Gradual Learning}
\author{Jan Feyereisl, Mat\v{e}j Nikl, Martin Poliak, Martin Str\'{a}nsk\'{y}, Michal Vlas\'{a}k\\ 
GoodAI, Prague, Czech Republic\thanks{Correspondence: \{jan.feyereisl, matej.nikl, martin.poliak, martin.stransky, martin.vlasak\}@goodai.com}}

\begin{document}

\maketitle

\begin{abstract}
  The General AI Challenge is an initiative to encourage the wider artificial intelligence community to focus on important problems in building intelligent machines with more general scope than is currently possible. The challenge comprises of multiple rounds, with the first round focusing on \emph{gradual learning}, i.e. the ability to re-use already learned knowledge for efficiently learning to solve subsequent problems. In this article, we will present details of the first round of the challenge, its inspiration and aims. We also outline a more formal description of the challenge and present a preliminary analysis of its curriculum, based on ideas from computational mechanics. We believe, that such formalism will allow for a more principled approach towards investigating tasks in the challenge, building new curricula and for potentially improving consequent challenge rounds.
\end{abstract}

\section{Introduction}
Existing artificial intelligence (AI) algorithms available today constitute the so-called narrow AI landscape, meaning that they have been designed, trained, and optimized by human engineers to solve a single, specific task or a very narrow collection of closely related problems. Although such algorithms sometimes outperform humans in their established skill-set, they are not able to extend their capabilities to new domains. This limits their re-usability, potentially increases the amount of data required to train them, and leaves them lacking generality and extensibility to higher order reasoning.

In contrast, algorithms capable of overcoming these limitations could eventually converge towards a repertoire of functionalities akin to a human-level skill-set. Such algorithms might be able to learn to come up with creative solutions for a wide range of multi-domain tasks. Systems employing aforementioned algorithms, oftentimes termed under the umbrella of general AI systems, are viewed by many as the ultimate leverage in solving many of humanity’s direst problems \cite{Bostrom2014,Domingos,Stone2016}.

\begin{figure}[!ht]
\centering
\includegraphics[clip, trim=4.1cm 10cm 2.5cm 6cm, width=0.48\textwidth]{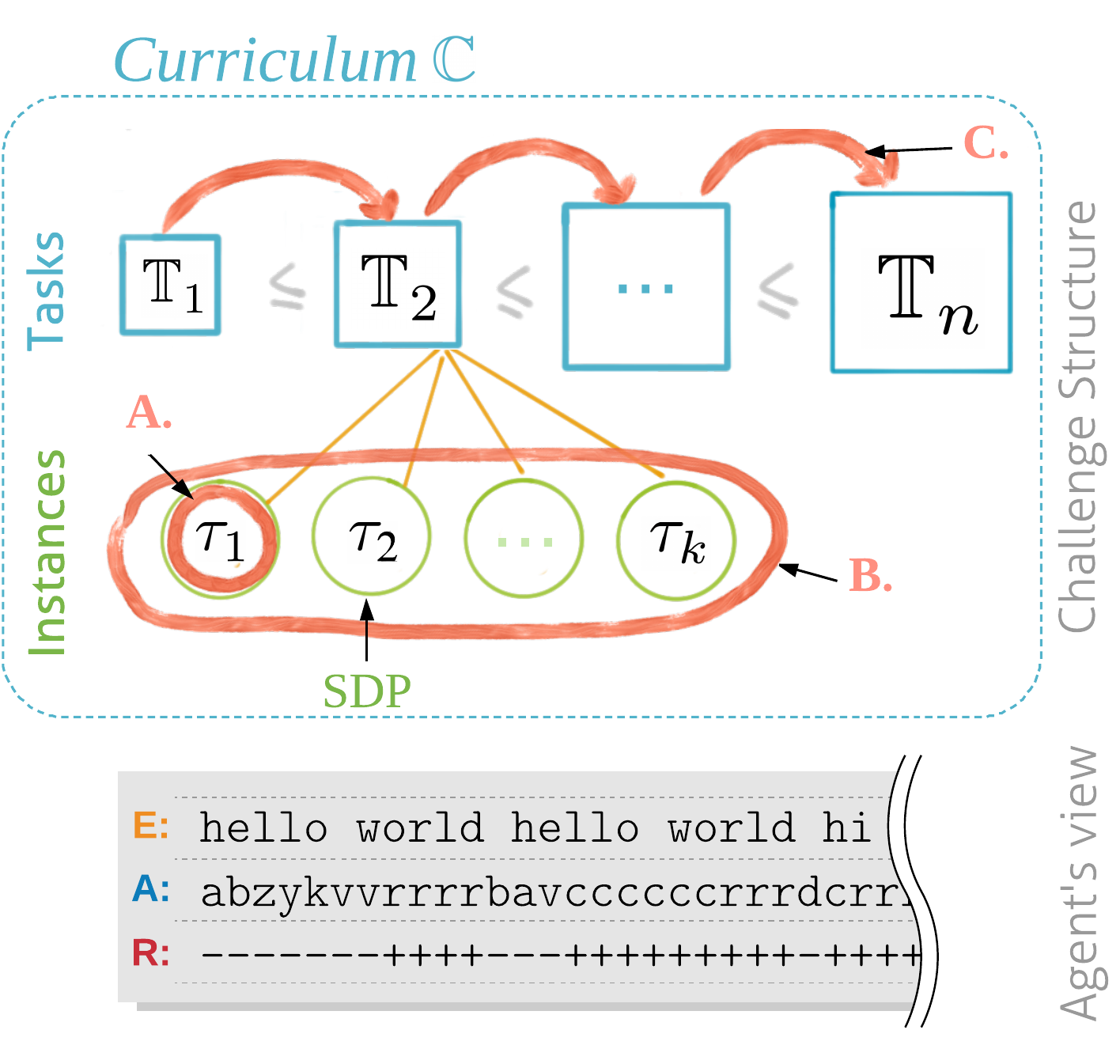}
\caption{
{\small Overview of the first round of the General AI Challenge. The structure of the challenge is hierarchical and multi-objective, and requires learning at multiple levels to solve a curriculum $\mathbb{C}$. From fast learning (\textbf{\textcolor{lred}{A.}}) at individual instance level of each task, where a `good enough' policy $\pi_i$ for a stochastic decision process (SDP) must be found, through shared sub-goal discovery (\textbf{\textcolor{lred}{B.}}) across instances $\tau_1,\dots,\tau_k$ leading to one-shot or few-shot learning, all the way to lower level knowledge and policy transfer (\textbf{\textcolor{lred}{C.}}) across different tasks $\mathbb{T}_i$ for speeding up convergence towards solutions on yet unseen tasks. What agent sees is shown at the bottom, i.e. continual streams of data with no instance or task separation information, where \textbf{\textcolor{env}{E}} denotes the environment stream, \textbf{\textcolor{agent}{A}} the agent's response and \textbf{\textcolor{reward}{R}} the reward.}
}
\label{fig:challenge}
\end{figure}

Gradual learning is an ability of learning systems to learn in a gradual manner. This is likely a necessary pre-condition for acquiring such a broad set of skills, striving to solve a wide range of disparate problems \cite{Rosa2016c,Kumaran2016a,Tommasino2016,Rusu2016-pp,Balcan2015,Pentina2015,Ruvolo2013,Hamker2001a}. 
Unfortunately, most existing algorithms struggle to deal with many tasks with different distributions at the same time and tasks that drift in distribution from one another~\cite{Ditzler2015} and frequently exhibit catastrophic forgetting, once applied to new problems, especially under realistic computational constraints~\cite{Rusu2016-pp,Fernando2017,Hamker2001a,Kirkpatrick2016,French1999,Gershman2015}.
Gradual learning is therefore a difficult and unsolved problem that requires a focused investigation by the community. The General AI Challenge provides a platform for such directed focus.

\section{Round One: Gradual Learning}
The aim of the first round of the challenge is to build an agent that exhibits \textbf{gradual learning}~\cite{Rosa2016c} as evidenced by solving a curriculum of increasingly complex and disparate sets of tasks. 
Despite the similarity in name to gradual learning in~\cite{Kumaran2016a} and other related concepts such as cumulative~\cite{Tommasino2016}, incremental~\cite{Rusu2016-pp}, life-long~\cite{Balcan2015,Pentina2015,Ruvolo2013,Hamker2001a} or continual learning~\cite{Kirkpatrick2016}, in our setting, the concept encompasses a broader class of requirements that an agent needs to satisfy. These include:

\begin{enumerate}
\item \textit{Exploiting previously learned knowledge}
\item \textit{Fast adaptation to unseen problems}
\item \textit{Avoiding catastrophic forgetting}
\end{enumerate}

The above requirements are expressed in two primary objectives of the first round, each with its own set of evaluation criteria:
\begin{myobj}
(Quantitative) {\normalfont Agent capable of passing an evaluation curriculum in the \emph{shortest number of simulation steps}. Passing requires the ability to exploit previously learned knowledge and avoid catastrophic forgetting, all within a predefined time limit.}
\end{myobj}
Alternatively, a description of a conceptual method for achieving gradual learning is also sought:
\begin{myobj}
(Qualitative) {\normalfont An idea, concept, or design that shows the best promise for scalable gradual learning. A working AI agent is desirable but not necessary.}
\end{myobj}

The fundamental focus of the challenge is on gradual learning, i.e.\ on the efficient re-use of already gained knowledge. Competitors are required to develop solutions with the following properties:

\begin{myprop}
{\normalfont(Gradual Learning)} An ability to re-use previously gained knowledge for the acquisition of subsequent knowledge to more efficiently solve hitherto new and yet unseen problems, while avoiding catastrophic forgetting.
\end{myprop}

It is important to note here not only the re-use of existing knowledge, but also the focus on \textit{efficiency} of solving new and unseen problems. Unlike compositionality or other concepts similar to gradual learning, our definition transcends the boundaries of meta-learning~\cite{Duan2017,Wang2016,Duan2017a,Santoro2016,Chen2016,Chen2016a}. We seek agents that are able to learn to \textit{quickly adapt} to new and unseen tasks, while exploiting the gradual structure of the underlying problems they are solving. One can think of the above as that the agents need to possess the ability to search for more efficient strategies to solve new problems.

Naturally, when solving new problems, agents must be able to retain the ability to solve old tasks. In many systems, the lack of such ability causes catastrophic forgetting~\cite{French1999}, which must be avoided:

\begin{myprop}
{\normalfont(Avoiding Catastrophic Forgetting)} Avoiding the loss of information, relevant for one task, due to the incorporation of knowledge necessary for a new task. 
\end{myprop}

In summary, the aim is not optimizing for agent’s performance of existing skills, i.e.\ how good an agent is at delivering solutions to problems it has already encountered. Instead, we desire optimizing for agent’s performance on solving new and unseen problems, i.e.\ maximizing the speed of convergence to `acceptable solutions' on new problems, while exploiting existing knowledge and ensuring its survival during the acquisition of new information.

\section{Background}
To fully appreciate our setting, below we provide background on why graduality can be beneficial and how it can be encouraged. This is followed by a more formal description of the challenge requirements, environment and evaluation procedures.

\subsection{Benefits of Graduality}
Given a complex task that needs to be solved, frequently a good strategy for finding a solution is to break the problem down into smaller problems which are easier to deal with. The same applies to learning~\cite{Bengio2009-fd,Salakhutdinov2013-on}. It can be much faster to learn things gradually than to try to learn a complex skill from scratch~\cite{Alexander2015-mx,Zaremba2015-ia,Gulcehre2016-mb,Oquab2014-ql}. 
One example of this is the hierarchical decomposition of a task into subtasks and the gradual learning of skills necessary for solving each of them~\cite{Krueger2009-vv,Vezhnevets2017,Lee2017,Andreas2017}, progressing from the bottom of the hierarchy to the top~\cite{Mhaskar2016-vq,Mhaskar2016-du,Poggio2015-hg,Polya2014-vb}. A prime example in the natural world is the gradual acquisition of motor skills by infants during their first years of life~\cite{Adolph2017}.

\subsection{Guidance through Curricula}
Exploiting graduality during learning is clearly beneficial. Building systems that learn in a gradual manner should then be encouraged and its benefits and limitations explored further. 
To enable such type of learning, one can control and guide the learning process. 
\textit{Guided learning}, also called curriculum learning,~\cite{Gulcehre2016-ht,Bengio2009-fd,Vapnik2015a} provides control by means of presenting the learner with parts of the problem in the order and extent that is likely to be most beneficial at that point in time. One method of providing such order is in the form of a learning curriculum~\cite{Bengio2009-fd}, akin to a curriculum used in schools. In this scenario, easier topics are taught before more complex ones, in order to exploit the gradual nature of taught knowledge.

Gradual and guided learning has a number of other benefits over other types of learning~\cite{Gulcehre2016-mb,Gulcehre2016-ht,Bengio2009-fd,Pan2010-en}.
For example, optimizing a model that has few parameters and gradually building up to a model with many parameters could be more efficient than starting with a model that has many parameters from the beginning~\cite{Stanley2002-nm}. In this case, a smaller number of new parameters is learned at each step~\cite{Chen2015-py,Kirkpatrick2016,Andreas2015-av,Rusu2016-pp}. This might also result in reducing the necessity for exploration.
Furthermore, apriori knowledge of the system's architecture might not be necessary~\cite{Zhou2012-ve,Rusu2016-pp,Ganegedara2016-cv}, and architecture size can be dynamically derived during training to correspond to the complexity of given problems~\cite{Fahlman1990-sp}.
Last but not least, reuse of already learned skills is feasible and encouraged~\cite{Andreas2015-av,Andreas2016-at}. 
Once a skill is acquired, it is no longer relevant how long the skill took to discover. The cost of using an existing skill is notably smaller than searching for a skill from scratch.

\section{Learning to Gradually Learn}
Having described the objectives of the challenge and established the benefits of gradual learning through a curriculum, we will now focus on formal definitions and descriptions of the requirements of the first round of the challenge.

\subsection{Instances, Tasks \& Curricula}
The gradual learning ability of an agent is evaluated by subjecting an agent to a sequence of $n$ tasks $\mathbb{T}_{1}, \mathbb{T}_{2}, \dots, \mathbb{T}_{n}$ of increasing complexity. Such sequence is called a curriculum:

\begin{mydef}
{\normalfont(Curriculum)} A curriculum $\mathbb{C} = ( \mathbb{T}_{1}, \dots, \mathbb{T}_{n} )$ is an n-tuple of tasks of increasing complexity. Tasks are endowed with an arbitrary measure of complexity.
\end{mydef}

In section \ref{sec:complexity}, we present one possible way of measuring task complexity in a principled manner and to ensure proper ordering of curricula. It can be performed with the help of a measure from the field of computational mechanics, namely statistical complexity $C_\mu$~\cite{Crutchfield1994}.

Each task is observed by an agent through task instances $ \tau \sim \mathbb{T}$. The publicly available curriculum provided as part of the challenge~\cite{Stransky2017} can be seen as a curriculum of distributions over stochastic decision problems (SDPs). In particular, a variant of partially observable Markov decision processes (POMDPs), or more generally partially observable stochastic games (POSGs).

\begin{mydef}
{\normalfont(Task and Instance)} A task $\mathbb{T}$ is a distribution over a set of Partially Observable Markov Decision Processes. An instance of a task $ \tau \sim \mathbb{T}$ is a sample from said distribution.
\end{mydef}

In other words, a single instance $\tau$ of a task $\mb{T}$ is a POMDP. A POMDP is an 8-tuple $(\mc{S}, \mc{A}, \Omega, \mc{P}, \mc{O}, r, \gamma, T)$ where $\mc{S}$ is a set of states, $\mc{A}$ a set of actions, $\Omega$ a set of observations, $\mc{P}: \mc{S} \times \mc{A} \times \mc{S} \to [0,1]$ a transition probability distribution, $\mc{O}: \mc{S} \times \mc{A} \times \Omega \to [0,1]$ is an observation probability distribution, $r:\mc{S} \times \mc{A} \to \mb{R}$ a reward function, $\gamma$ a discount factor and $T$ a horizon. In the standard POMDP formulation, the goal of an agent is to maximize its future discounted reward $\mb{E}_{\sigma} \left[ \sum_{t=0}^{T}\gamma^t r(s_t,a_t) \right]$, where the expectation is over the sequence of agent's belief state/action pairs $\sigma = ((b_0,a_0),\dots,(b_T,a_T))$, where $b_0(s) = \mathbb{P}(\mc{S} = s)$ is an initial belief state, $a_t \sim \pi_{\theta}(a_t|b_t)$, $b_{t+1} = \mathbb{P}(b_{t+1}|b_t, a_t, o_t)$ and $\pi_\theta$ denotes a policy parameterized by $\theta$.

Unlike in the standard setting however, the goal in the challenge round is different and distributed across multiple levels of hierarchy. We are not interested in maximizing the agent's future discounted reward, but rather a more complex set of objectives at different scales. For example at the instance level, it is sufficient to find an acceptable solution to each instance $\tau_i$, as described in Algorithm \ref{alg:sic}. Figure \ref{fig:challenge} shows the three levels of hierarchy present in the challenge:
\begin{enumerate}
\item \textbf{Fast learning} (\textbf{\textcolor{lred}{A}}): \textit{Policy Search} $-$ At the individual instance level, the agent is required to find a solution (e.g.\ a policy $\pi$) to a single POMDP/task instance $\tau$.
\item \textbf{Slow learning} (\textbf{\textcolor{lred}{B}}): \textit{Meta-Policy Discovery} $-$ The agent needs to discover a meta-strategy (e.g. a meta-policy $\pi^{meta}$) that quickly converges to a solution across all instances $\tau_i$.
\item \textbf{Fast Adaptation} (\textbf{\textcolor{lred}{C}}): \textit{Policy Transfer} $-$ The agent is required to exploit existing acquired knowledge and policies in order to adapt to each new task in a curriculum, to solve it faster.
\end{enumerate}
Whether it is possible to find a correspondence between the challenge objectives and the standard reward formulation remains to be seen and up to competitors to determine.

In addition to the above hierarchy, a number of objectives and constraints need to be satisfied by competing agents. 
The primary goal of an agent is the completion of the entire curricula $\mathbb{C}$ in as short a time as possible. 

\begin{mydef}
{\normalfont(Quantitative Objective)} Successful completion of an evaluation curriculum $\mathbb{C}_e$ in the shortest number of time-steps possible among all competitors and within 24 hours from the start of the evaluation process on predefined evaluation H/W. Given a set of competing agents $A$, the fastest agent is determined according to:
\[
\argmin_{\alpha \in A} \varrho(\alpha, \mathbb{C}_e)
\]
where $\varrho: A \times \mathbb{C} \to \mathbb{N}_+$ corresponds to $\mathtt{RunCurriculum}()$ in Algorithm \ref{alg:ptc} which returns the number of simulation steps it took an agent $\alpha$ to successfully complete a curriculum $\mathbb{C}$.
\label{def:obj}
\end{mydef}

Satisfying the condition set forth in Definition \ref{def:obj}, the agent also has to show that two additional conditions are met, namely gradual learning and avoiding catastrophic forgetting.

\begin{mydef}
{\normalfont(Gradual Learning)} A manifestation of Property 1, exhibited through a reduced number of computational steps required when solving a task $\mathbb{T}$ ensuing the solving of previous tasks, i.e.\ $\varrho(\alpha_{(\mathbb{T}_1,\mathbb{T}_2)}, (\mathbb{T}_3)) < \varrho(\alpha, (\mathbb{T}_3))$, where $\alpha_{(\mathbb{T}_1,\mathbb{T}_2)}$ denotes agent $\alpha$ that has already learned to solve tasks $\mathbb{T}_1$ and $\mathbb{T}_2$.
\end{mydef}

The above condition does not ensure that gradual learning truly occurs, nevertheless it provides sufficient evidence that an agent improves its performance on subsequent tasks, having already solved other tasks before.

\begin{mydef}
{\normalfont(Avoiding Catastrophic Forgetting)} A manifestation of Property 2 through the maintenance of fast convergence to an acceptable solution for already solved tasks, i.e.\ $\varrho(\alpha_{(\mathbb{T}_1,\mathbb{T}_2,\mathbb{T}_3)}, (\mathbb{T}_2)) \leq c \varrho(\alpha_{(\mathbb{T}_1,\mathbb{T}_2)}, (\mathbb{T}_2))$ where $c$ is some constant of the agent.
\end{mydef}

This condition ensures that learning to solve a new task does not impair the ability to solve previously encountered tasks.

\subsection{Validation Curricula}
To test generalization of gradual learning, an agent trained on a single curriculum $\mathbb{C}$ cannot be effectively evaluated on tasks from the same curriculum. A different curriculum is necessary. 
Curricula appropriate for a gradually learning agent also form a distribution $\mathfrak{C}$ from which a training and an evaluation curriculum should be drawn. The mini and micro tasks used in the challenge (described in section \ref{sec:env}) form together one sample from this distribution. Using a single sample curriculum $\mathbb{C} \sim \mathfrak{C}$ to estimate the distribution $\mathfrak{C}$ might be too difficult. It is possible that competitors might need to create their own curricula. Examples of a number of possible final tasks from alternative curricula were shown in \cite{Rosa2017}.

\section{Evaluation}
As mentioned previously, simply maximizing reward $R$ is neither sufficient, nor desired. Immediately after the agent reaches an acceptable performance $R^*$ on an instance, the environment presents it with the next sample $\tau$. Upon the successful completion of a number of instances from the same task, determined according to Algorithm \ref{alg:sic}, the environment presents the agent with the next task in the curriculum. This can be seen in Algorithm \ref{alg:ptc}.

\begin{algorithm}[ht]
  \SetKwProg{Function}{function}{}{end}
  \SetKwFunction{RunInstance}{Solve}
  \SetKwFunction{CalcHardLimit}{HardLimit}
  \SetKwFunction{AgentStep}{AgentStep}
  \SetKwFunction{EnvStep}{EnvStep}
  
  \SetKwData{timeStep}{$t$}
  \SetKwData{reqRewardInRow}{$R^{*}$}
  \SetKwData{rewardInRow}{$R_{+}$}
  \SetKwData{instance}{$\tau_{i}$}
  \SetKwData{agent}{$\alpha$}
  \SetKwData{reward}{$R$}
  \SetKwData{input}{$o_t$}
  \SetKwData{output}{$a_t$}
  
  \SetKw{Or}{or}
  \SetKw{Return}{return}
  
  \SetKwInOut{Input}{input}
  \SetKwInOut{Output}{output}
  
  \Input{An agent \agent, a task instance \instance, the minimum number of positive rewards \reqRewardInRow, uninterrupted by a negative one, required to solve current instance}  
  \Output{The number of steps \timeStep taken to solve \instance.}
  \BlankLine
  
  \Function{\RunInstance{\agent, \instance, \reqRewardInRow}}
  {
    \timeStep $\larr 0$; \hfill \textcolor{mygray}{{\small \# Counter of elapsed steps}} \\
    \rewardInRow $\larr 0$; \hfill \textcolor{mygray}{{\small \# Counter of positive rewards}} \\
    \reward, \input, \instance $\larr$ \EnvStep{\instance, $\varnothing$}; 
	\hfill \textcolor{mygray}{{\small \# Environment}} \\
    \BlankLine
    
    \textcolor{mygray2}{{\small \# Agent-environment loop until reward/time-limit reached}} \\
    \Repeat{\rewardInRow $=$ \reqRewardInRow \Or \timeStep $=$ \CalcHardLimit{\instance}}
    {
      \agent, \output $\larr$ \AgentStep{\agent, \reward, \input}; 
      \hfill \textcolor{mygray}{{\small \# Agent}} \\
      \reward, \input, \instance $\larr$ \EnvStep{\instance, \output};
      \hfill \textcolor{mygray}{{\small \# Environment}} \\
      \timeStep $\larr \timeStep + 1$; \hfill \textcolor{mygray}{{\small \# Increment elapsed steps}} \\
      \BlankLine
      
      \textcolor{mygray2}{{\small \# If correct, increment reward counter, else re-set it}} \\
      \uIf{\reward $= 1$}
      {
        \rewardInRow $\larr \rewardInRow +1$; \hfill \textcolor{mygray}{{\small \# Increment reward counter}} \\
      }
      \ElseIf{\reward $= -1$}
      {
      	\rewardInRow $\larr 0$; \hfill \textcolor{mygray}{{\small \# Reset positive reward counter}} \\
      }
    }
    
    \Return \timeStep; \hfill \textcolor{mygray}{{\small \# No. of steps taken to solve instance \instance}} \\
  }
  
  \caption{Progression through one task instance $\tau$}
  \label{alg:sic}
\end{algorithm}

The \texttt{EnvStep} and \texttt{AgentStep} functions in Algorithm \ref{alg:sic} update the environment and the agent respectively, while exchanging reward, input and output. Together, they form the core of the environment-agent communication loop, depicted in Figure \ref{fig:env}. 

The helper functions \texttt{SoftLimit} and \texttt{HardLimit} are used to compute the respective time step limits for a given instance.
As the naming suggests, there are two types of limits considered when running an instance: a soft limit, and a hard limit.
Both limits are represented in terms of a number of environment steps. They are dynamically computed as they depend on the current instance $-$ in fact, they can even evolve as the agent progresses through the instance.

The soft limit can be thought of as a \textit{success} limit. The agent is required to solve the instance within this limit in order for this attempt to count as successful.
If it fails to do so, the prospective solution will no longer be considered as successful. The instance, however, does not end yet. This allows the agent to continue exploring and gather some additional useful knowledge about the unsolved problem.

A forceful instance termination comes only when the \textit{hard} limit is reached. This behavior is beneficial in situations where an agent gets stuck in a particular instance.

The reasoning behind such limits is as follows: the agent has no way of communicating its needs, for example the need to change the instance to a different one, or switching to the previous task. Therefore, those limits can be thought of as supplement to such cases.

\begin{algorithm}[ht]
  \SetKwProg{Function}{function}{}{end}
  \SetKwFunction{RunCurriculum}{RunCurriculum}
  \SetKwFunction{SampleInstance}{SampleInstance}
  \SetKwFunction{GetReqReward}{ReqReward}
  \SetKwFunction{RunInstance}{Solve}
  \SetKwFunction{CalcSoftLimit}{SoftLimit}
  
  \SetKwData{agent}{$\alpha$}
  \SetKwData{totalTimeSteps}{$T$}
  \SetKwData{curriculum}{$\mathbb{C}$}
  \SetKwData{reqSuccInRow}{$N_{s}$}
  \SetKwData{task}{$\mathbb{T}$}
  \SetKwData{succInRow}{$N_{j}$}
  \SetKwData{reqReward}{$R_{j}$}
  \SetKwData{instance}{$\tau_{i}$}
  \SetKwData{timeSteps}{$t_i$}
  \SetKwData{softLimit}{$t_s$}

  \SetKw{In}{in}
  \SetKw{Return}{return}
  
  \SetKwInOut{Input}{input}
  \SetKwInOut{Output}{output}
 
  \Input{An agent \agent, a curriculum \curriculum and the number of successfully solved consecutive instances \reqSuccInRow required}
  \Output{The total number of steps \totalTimeSteps the agent needed to solve curriculum \curriculum}
  \BlankLine
  
  \Function{\RunCurriculum{\agent, \curriculum, \reqSuccInRow}}{
    \totalTimeSteps $\larr 0$; \hfill \textcolor{mygray}{{\small \# Counter of elapsed steps}} \\
    
    \textcolor{mygray2}{{\small \# Solve each task $\mathbb{T}$ in a curriculum \curriculum}} \\
    
    \ForEach{\task \In \curriculum}{
      \succInRow $\larr$ 0; \hfill \textcolor{mygray}{{\small \# Counter of consecutive successes}} \\
      \reqReward $\larr$ \GetReqReward{\task}; \hfill \textcolor{mygray}{{\small \# Required min. reward}} \\
      \textcolor{mygray2}{{\small \# Successfully solve \reqSuccInRow instances of task $\mathbb{T}$}} \\
      \Repeat{\succInRow $=$ \reqSuccInRow}{
        \instance $\sim$ \task; \hfill \textcolor{mygray}{{\small \# Sample a new instance}} \\        
        \timeSteps $\larr$ \RunInstance{\agent, \instance, \reqReward}; \hfill \textcolor{mygray}{{\small \# Solve instance}} \\
        \totalTimeSteps $\larr \totalTimeSteps + \timeSteps$; \hfill \textcolor{mygray}{{\small \# Increment elapsed steps}} \\
        \softLimit $\larr$ \CalcSoftLimit{\instance}; \hfill \textcolor{mygray}{{\small \# Calc. soft limit}} \\
        \BlankLine
		\textcolor{mygray2}{{\small \# If solved sufficiently quickly}} \\
        \uIf{\timeSteps $\le$ \softLimit}
        {
          \succInRow $\larr \succInRow + 1$; \hfill \textcolor{mygray}{{\small \# increment success counter}} \\
        }
        \Else
        {
          \succInRow $\larr 0$; \hfill \textcolor{mygray}{{\small \# else reset counter}} \\
        }
      }
    }
    
    \Return \totalTimeSteps; \hfill \textcolor{mygray}{{\small \# No. of steps taken to solve \curriculum}} \\
  }
  
  \caption{Progression through a curriculum $\mathbb{C}$}
  \label{alg:ptc}
\end{algorithm}

\section{Learning Environment}
\label{sec:env}
Participants are provided with an environment and a public curriculum of tasks of increasing complexity~\cite{Stransky2017}. Both, the environment and the associated tasks are specifically constructed to minimize distraction from unrelated research problems and to primarily focus on gradual learning and associated obstacles. The environment and tasks are built on top of the CommAI-env~\cite{Mikolov2015,Baroni2017}.

Participants are asked to develop and train their agents on the public curriculum and possibly any other additional knowledge or data sources that they have at their disposal.
The public set of tasks provides a reference example of how the evaluation curriculum might be structured and the type of tasks that will be required to be solved by an agent.
Tasks come in two groups, as simple \textit{micro-tasks} and as more advanced \textit{mini-tasks}. It is important to note, that natural language processing is not necessary to solve the tasks.

\subsection{Mini-tasks}
Mini-tasks are based directly on the CommAI-mini task set~\cite{Baroni2017}, with focus on simple grammar processing and deciding language membership problems for strings. The mini-tasks are simple for an educated and biased human, but they are tremendously complex for an unbiased agent that receives only sparse reward. For this reason, we believe that simpler tasks are necessary. We refer to those as \textit{micro-tasks}. 

\subsection{Micro-tasks}
The purpose of micro-tasks is to allow the agent to acquire a sufficient repertoire of prior knowledge, necessary for discovering the solutions to the more complex mini-tasks in a reasonable time, under the assumption that the agent is capable of gradual learning.

Decomposing mini-tasks into simpler problems yields a number of skills that the agent should learn first. For each such skill, there is a separate micro-task. The micro-tasks build on top of each other in a gradual and compositional way, relying on the assumption that the agent can make use of them efficiently.

One could argue that the micro-tasks (and even the mini-tasks) are too simple and can be solved by hard-coding the necessary knowledge into the agent. This is indeed true. However, such a solution is not desirable and goes against the spirit of the challenge - to learn new skills from data. The challenge will prevent any hard-coded solution by using hidden evaluation tasks which are different from the public curriculum. A hard-coded solution that does not exhibit gradual learning should fail on such evaluation tasks.

\subsection{Environment-Agent Interface}

\begin{figure}
\centering
\includegraphics[clip, trim=7.3cm 7.2cm 7.6cm 5.7cm, width=0.50\textwidth]{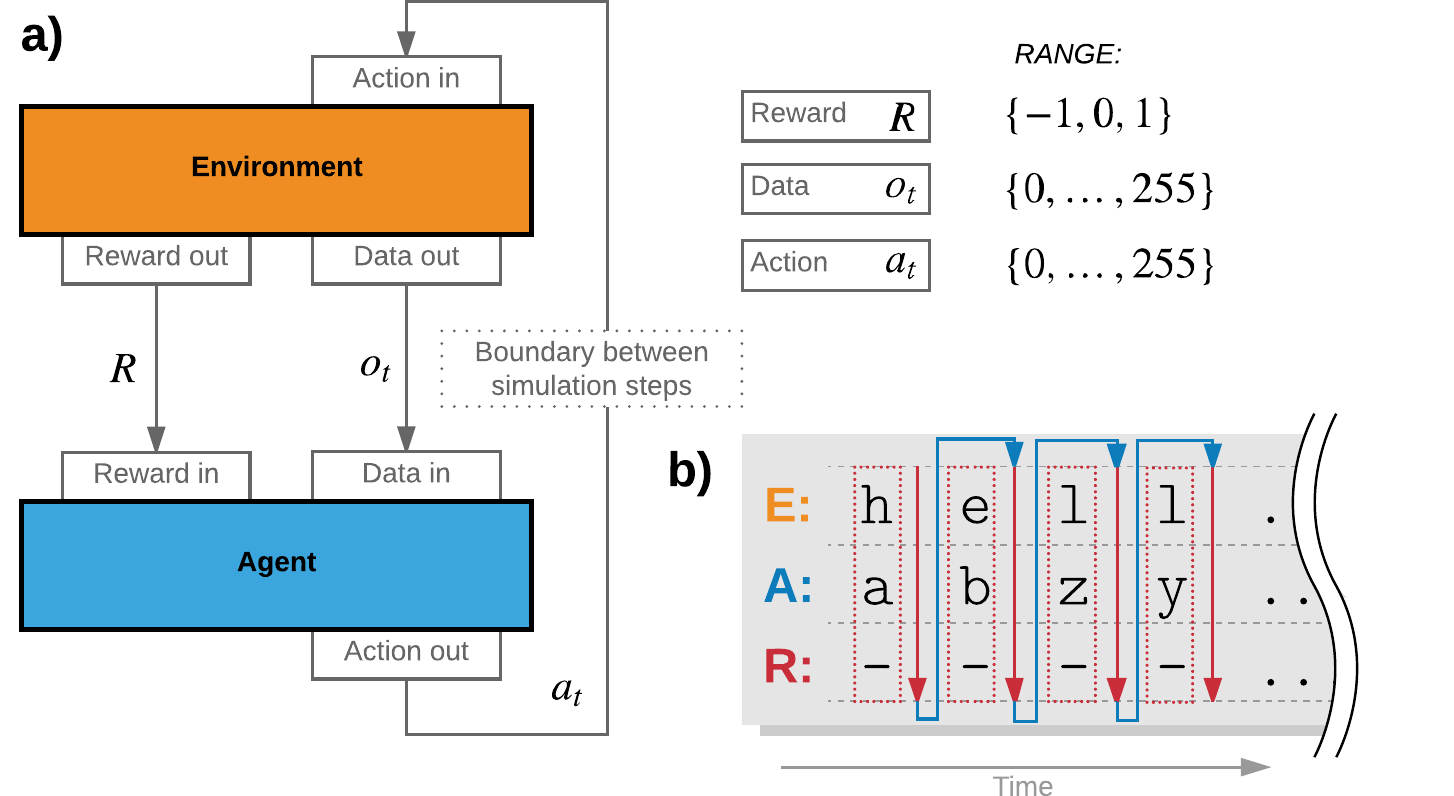}
\caption{a) Interface between the agent and the environment. b) Depiction of environment output with overlaid simulation order.}
\label{fig:env}
\end{figure}

The interface between the agent and the environment is depicted in Figure \ref{fig:env}.
The environment sends reward $\{-1,0,1\}$ and data (one byte) to the agent and receives the agent’s action (one byte) in response. This happens in a continuous cycle. During a single simulation step, the environment processes the received action from the agent and sends reward with new data to the agent; the agent processes this input and sends an action back to the environment.

Specifically, the environment sends 8 bits of data to the agent at every simulation step and receives 8 bits of data back from the agent. This is different from the original CommAI-env~\cite{Mikolov2015}, which sends only a single bit instead of a full byte every time. Although this makes the interface slightly less flexible, the agent’s communication should be more interpretable and it should reduce the complexity of the presented problems.

Note that the environment is not sending any special information about what task the agent is in at any point in time. The environment appears as a continuous stream of bytes (and rewards) to the agent.

\section{Discussion}
Looking back at Figure \ref{fig:challenge}, it is apparent that there is a significant gap between the structure of the curriculum, the objectives to be learned within, and what kind of information the agent receives. Most notably, there is no explicit information about a successful completion of an instance or a task. Similarly, no indication of an instance or a task switch is provided. The lack of information and limited amount of feedback that an agent has at its disposal makes this round challenging and different from most other learning problems.

Moreover, unlike standard POMDPs, for many of the tasks in the public curriculum, the reward is adaptive. This can be very challenging due to the drifting nature of the reward function.
Another challenging aspect of the first round is the inability to request previous tasks from the environment. Hence, competing agents need to be sample efficient as well as `epoch' efficient.

\section{Task Complexity via Computational Mechanics}
Computational mechanics \cite{Crutchfield2013}, a subfield of physics, is concerned with exploring the ways in which nature performs computations. How to identify structure in natural processes and how to measure it, as well as how information processing is embedded in dynamical behavior.
The field offers a number of useful tools for thinking about and quantifying complex systems, some of which are relevant for building intelligent machines. Namely, we are interested in the concept of $\epsilon$-machines~\cite{Crutchfield1989} and the associated measure of statistical complexity $C_{\mu}$.

In our scenario, these concepts and tools can be useful for exploring curricula of the challenge, from the point of view of the complexity of each individual task and the ability to begin investigating the possibility of automating the creation of more principled curricula for both training and evaluation.

\subsection{$\epsilon$-Machines as Minimal Optimal Models}
The concept of $\epsilon$-machines as a class of minimal optimal predictors was developed for quantifying structure in a stochastic process via the reconstruction of underlying causal states in a natural process \cite{Crutchfield1994}. In their most common form, they can be thought of as a type of hidden Markov model (HMM), whose states have a unique, causal definition. $\epsilon$-machines are, however, not limited to only a HMM representation and can in fact be used at various levels of representation hierarchy, depending on whether current representation hits the limits of the agent's computational resources \cite{Crutchfield1994}.

An $\epsilon$-machine can be constructed with the help of an equivalence relation
\begin{equation}
\past{y} \sime \past{y}' \Longleftrightarrow
\mb{P}\left( \future{Y}|\past{Y} = \past{y} \right)
= \mb{P}\left( \future{Y}|\past{Y} = \past{y}' \right) 
\label{eq:eqivalence}
\end{equation}
where $Y$ denotes a discrete random variable, $\past{Y}$ and $\future{Y}$ a block of past (e.g. $\past{Y_{t}} = \ldots Y_{t-2}Y_{t-1}$ ) and future random variables, respectively, with $\past{y}$ and $\future{y}$ denoting a particular history or a future (a sequence of symbols from alphabet $\bmc{Y}$) of a generating process, respectively. The equivalence relation (\ref{eq:eqivalence}) states that any two histories $\past{y}$ and $\past{y}'$ are equivalent if the probability distribution over their futures is the same, i.e. $\mb{P}\left( \future{Y}|\past{Y} = \past{y} \right) = \mb{P}\left( \future{Y}|\past{Y} = \past{y}' \right)$. 
Given the above equivalence relation, one can then define a mapping from the space of histories $\past{\bmc{Y}}$ to `causal' states of the underlying process. Such mapping $\epsilon : \past{\bmc{Y}} \rightarrow \bmc{S}$ is called an $\epsilon$-map that takes a given past $\past{y}$ to its corresponding causal state $\sigma_i$:
\begin{equation*}
\epsilon \left( \past{y} \right) = \sigma_i
= \left\{ \past{y}' : \past{y} \sime \past{y}' \right\}
\end{equation*}
Intuitively, one can think of $\epsilon$ as a function that partitions the set of pasts $\past{\bmc{Y}}$ according to the equivalence relation (\ref{eq:eqivalence}), into clusters that lead to the same distribution over futures.

We can then define the $\epsilon$-machine as a tuple $(\bmc{Y}, \bmc{S}, \bmc{T})$ where $\bmc{Y}$ denotes the process' alphabet, $\bmc{S}$ its causal state set, and $\bmc{T}$ a set of symbol-valued transition matrices:
\begin{equation*}
\bmc{T} \equiv \left\{ T^{(y)} \right\}_{y \in \bmc{Y}}
\end{equation*}
where $T^{(y)}$ comprises the following elements:
\begin{equation*}
T_{ij}^{(y)} = \mb{P}(S_1 = \sigma_j, Y_0 = y | S_0 = \sigma_i),
\end{equation*}
where $S_0$ and $S_1$ denote two temporally consecutive random variables of a random variable chain defining the causal state process that underlies the stochastic process we are modeling. Each $S_t$ takes on some value $s_t=\sigma_i \in \bmc{S}$. The set $\bmc{T}$ thus defines the dynamics over causal states of the underlying system.
An $\epsilon$-machine is then a unique, minimal-size, maximally predictive unifilar representation of the observed process \cite{Crutchfield1989}. 

Despite their desirable minimality and optimality properties, $\epsilon$-machines are inherently able to model generative processes only. In order to exploit the useful properties that $\epsilon$-machine's formulation offers for the analysis of our curriculum, we need to go beyond output processes and exploit a slightly more complex representation, that of input-output processes, namely $\epsilon$-transducers.

\subsection{$\epsilon$-Transducer}
In our scenario, we are interested in observing the structure of not only one process, but a coupling between two stochastic processes, which can also be viewed as the analysis of a communications channel between an agent and its environment. The concept of an equivalence relation can be extended, for this type of scenario, over joint (input-output) pasts:
\begin{multline}
\past{(x,y)} \sime \past{(x,y)}' \Longleftrightarrow \\
\mb{P}\left( \future{Y}|\future{X}, \past{(X,Y)} = \past{(x,y)} \right) \\
= \mb{P}\left( \future{Y}|\future{X}, \past{(X,Y)} = \past{(x,y)}' \right)
\end{multline},
defining a basis for a channel's unique, maximally predictive, minimal statistical complexity unifilar presentation, called $\epsilon$-transducer. Similarly to $\epsilon$-machines, a map from pasts to states can be created
$\epsilon : \past{(\bmc{X}, \bmc{Y})} \rightarrow \bmc{S}$ that maps given joint input-output past $\past{(x,y)}$ to its corresponding channel causal state
\begin{equation*}
\epsilon \left( \past{(x, y)} \right) = \sigma_i
= \left\{ \past{(x, y)}' : \past{(x, y)} \sime \past{(x, y)}' \right\}.
\end{equation*}
The $\epsilon$-transducer is then defined as the tuple $(\bmc{X}, \bmc{Y}, \bmc{S}, \bmc{T})$, where, in contrast to $\epsilon$-machines, $\bmc{X}$ additionally defines the set of inputs and $\bmc{T}$ the set of conditional transition probabilities:
\begin{equation*}
\bmc{T} \equiv \left\{ T^{(y|x)} \right\}_{x \in \bmc{X},y \in \bmc{Y}}
\end{equation*}
where $T^{(y|x)}$ has elements:
\begin{equation*}
T_{ij}^{(y|x)} = \mb{P}(S_1 = \sigma_j, Y_0 = y|S_0 = \sigma_i, X_0 = x)
\end{equation*}
The ability of $\epsilon$-transducers to model input-output mappings, enables the incorporation of actions and makes modeling of tasks in a curriculum $\mathbb{C}$ feasible.

\subsection{Statistical and Structural Complexity}
\label{sec:complexity}
Having defined both $\epsilon$-machines and $\epsilon$-transducers, we can now define 
a distribution $\tilde{\pi}$ over causal states for $\epsilon$-machines as
\begin{align*}
\tilde{\pi}(i) &= \mb{P}(S_0 = \sigma_i) \\
         &= \mb{P}\left( \epsilon \left( \past{Y} \right) = \sigma_i \right)
\end{align*}
and an input-dependent state distribution $\tilde{\pi}_X$ for $\epsilon$-transducers:
\begin{align*}
\tilde{\pi}_X(i) &= \mb{P}_X(S_0 = \sigma_i) \\
         &= \mb{P}_X\left( \epsilon \left( \past{(X,Y)} \right) = \sigma_i \right)
\end{align*}
In both cases, this defines the asymptotic probability of a process being in any one of its causal states. This information can then be used for quantifying the complexity of the underlying generating or input-output process, respectively. This measure, also called \textit{statistical complexity} in the case of $\epsilon$-machines, is defined as $C_\mu = \mathrm{H}[\tilde{\pi}]$, where $\mu$ points to the fact that it is a measure over an asymptotic distribution, and $\mathrm{H}$ denotes the Shannon entropy \cite{Shannon48}.  
On the other hand, $\epsilon$-transducer's input-dependent statistical complexity is defined as
\begin{equation*}
C_X = \mathrm{H}[\tilde{\pi}_X].
\end{equation*}
The upper bound on $\epsilon$-transducer complexity is then the channel complexity, calculated as the supremum of statistical complexity over input processes:
\begin{equation*}
\overline{C_\mu} = \sup_X C_X
\end{equation*}
with topological state complexity being $C_0 = \log_2 |\bmc{S}|$ and due to the fact that uniform distributions maximize Shannon entropy, in general $\overline{C_\mu} \leq C_0$ \cite{Barnett2015a}.

Statistical complexity $C_{\mu}$ is a widely used measure of complexity in physics and complexity science. It has a number of benefits over other measures and intuitive interpretations. For example, it measures the amount of information a process needs to store in order to be able to reconstruct a given signal, or in the case of structural complexity, it defines the capacity of a communication channel.
In the case of tasks in our challenge, it could provide a principled way of measuring an upper bound on task complexity that can be beneficial in the construction of structured curricula and subsequent automatic generation of tasks of increasing complexity.

\section{Agent-agnostic Task and Curriculum Representation}
Modeling tasks $\mathbb{T}$ of the challenge curriculum $\mathbb{C}$ with the help of $\epsilon$-transducers allow for an agent-agnostic representation of the complexity of the problem each task tackles. It not only provides a way to measure complexity of each task, but also to compare and contrast entire task curricula and reason about the inherent compositionality and graduality of an entire curriculum (see Figure (\ref{sfig:eps2})). Preliminary results for a number of tasks from the challenge curriculum can be seen in the appendix.

\subsection{Task and Curriculum Complexity}
Using statistical and structural complexity we are able to start reasoning about the correctness of the curriculum order. This allows for more well defined curricula and subsequent possibility of automating the curriculum generating process with guaranteed characteristics.

\subsection{Task and Curriculum Graduality}
The minimal representation of tasks also allows us to start thinking about comparison of how much structure is shared among tasks within a curriculum. This can be beneficial in designing and measuring the intrinsic graduality of a curriculum. As $\epsilon$-machines and transducers are inherently graphical models, graph theoretic tools and algorithms, such as subgraph-matching, can be used to start comparing tasks and measure their level of shared structure that can be eventually exploited by an agent with gradual learning capabilities.

\section{Conclusions}
In this work, we have outlined the structure and goals of the first round of the General AI Challenge, focused on building agents that can learn gradually. We have argued that building such agents requires solving problems at three different levels of hierarchy. From fast learning, akin to a search for a policy at a level of an instance of a task, through slow learning that requires the discovery of a meta-policy that generalizes across instance, all the way to fast adaptation to new tasks, akin to policy transfer between disparate problems. Furthermore, we have proposed the use of a method from computational mechanics, namely $\epsilon$-transducers for modeling of tasks in a curriculum that allow for quantifying task complexity and potentially graduality of entire curricula. Explicit calculation of the complexity of tasks and its subsequent use for building better curricula is left for future work.

\section*{Acknowledgements}
We would like to thank the entire GoodAI team, in particular Olga Afanasjeva and Marek Rosa.

\bibliographystyle{named}
\bibliography{biblio}

\onecolumn
\appendix
\section{Appendix}

\begin{sidewaysfigure}
\centering
\includegraphics[clip, trim=0cm 0cm 0cm 0cm, width=0.95\textwidth]{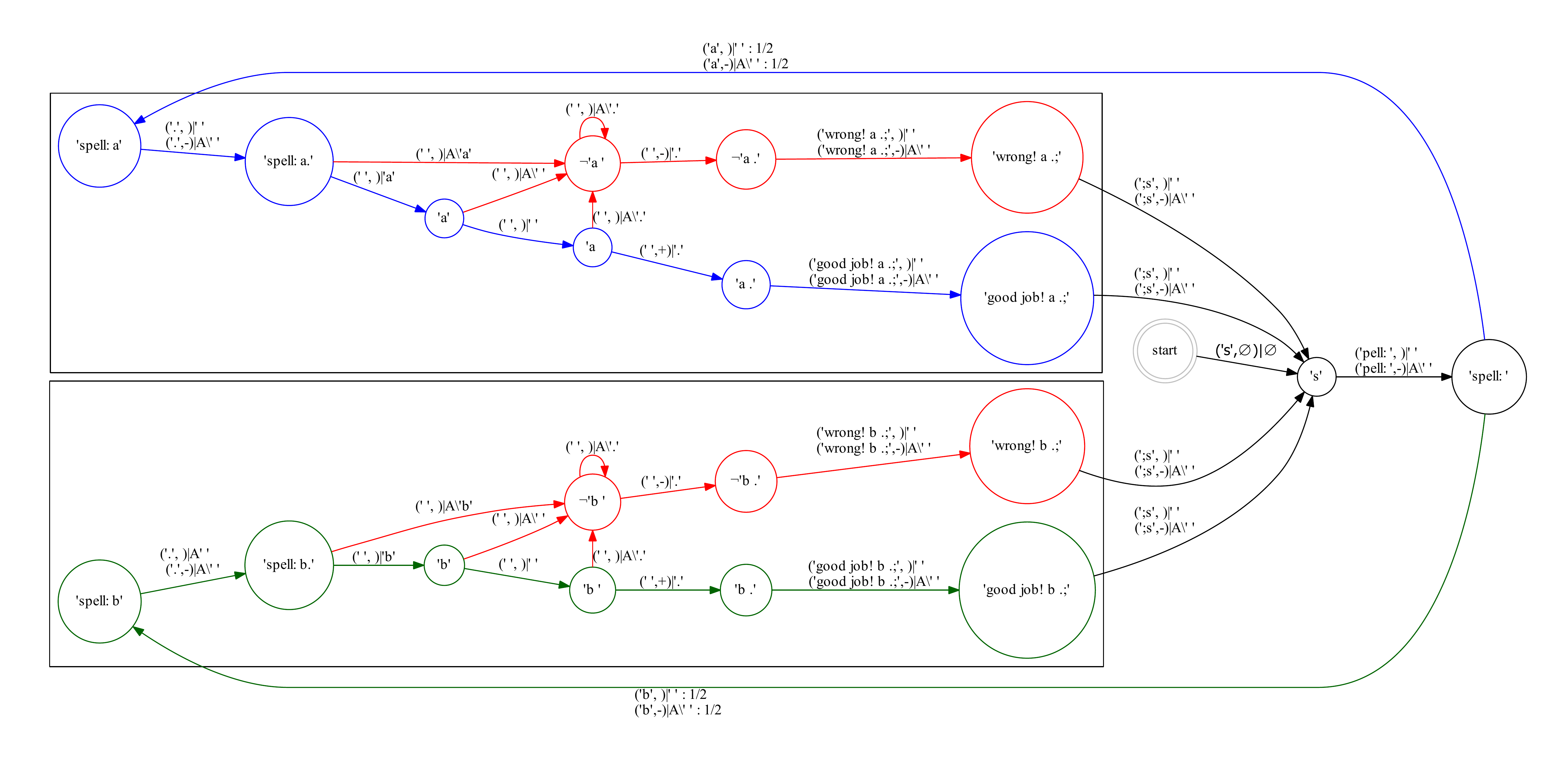}
\caption{A simplified visualization of $\epsilon$-transducer for task A.2.9.1 from the challenge curriculum.}
\label{fig:eps3}
\end{sidewaysfigure}

All figures in this section are simplified visualizations of $\epsilon$-transducers, each for a particular task. These simplifications come in four flavors:
\begin{enumerate}
  \item Only two instances, out of a numerous possible, are shown, as others would only be repetitions of an already presented pattern (black rectangles delineate instances).
  \item For the sake of clarity not all transitions (arrows) are always visualized (e.g.\ error or instance switch transitions).
  \item All transition probabilities from a state are uniformly distributed among visualized transitions. Only instance decision transition probabilities are shown for illustration.
  \item In Figure~\ref{fig:eps3}, states are merged together, to avoid repetition when the environment is providing the current instance description. This way the visualization is still readable and focuses more on the important part $-$ the part where the agent replies.
\end{enumerate}
General notation for transitions (arrows) between causal states is as follows: $(o_{t+1},r_{t}|a_t)$, where $o_t$ is an observation/state that is the output of the environment, $r_t$ the reward for the last performed action $a_{t-1}$ by the agent. Note that the action $a_t$ at current time-step determines the reward $r_t$ at current time-step, but the observation/state for the \textit{subsequent} step.
\begin{figure}[h]
\captionsetup[subfigure]{aboveskip=-15pt,belowskip=0pt}
\centering
\begin{subfigure}{\textwidth}
  \centering
  \includegraphics[clip, trim=0cm 0cm 0cm 0cm, width=0.9\textwidth]{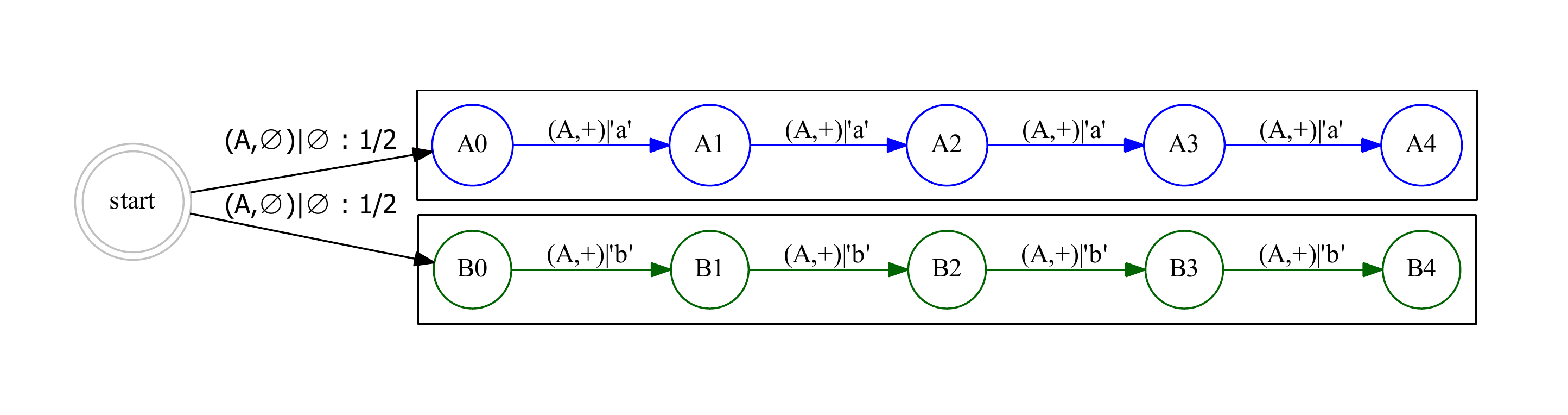}
  \caption{}
  \label{sfig:eps1}
\end{subfigure}
\vspace{-\baselineskip} \\
\begin{subfigure}{\textwidth}
  \centering
  \includegraphics[clip, trim=0cm 0cm 0cm 0cm, width=0.9\textwidth]{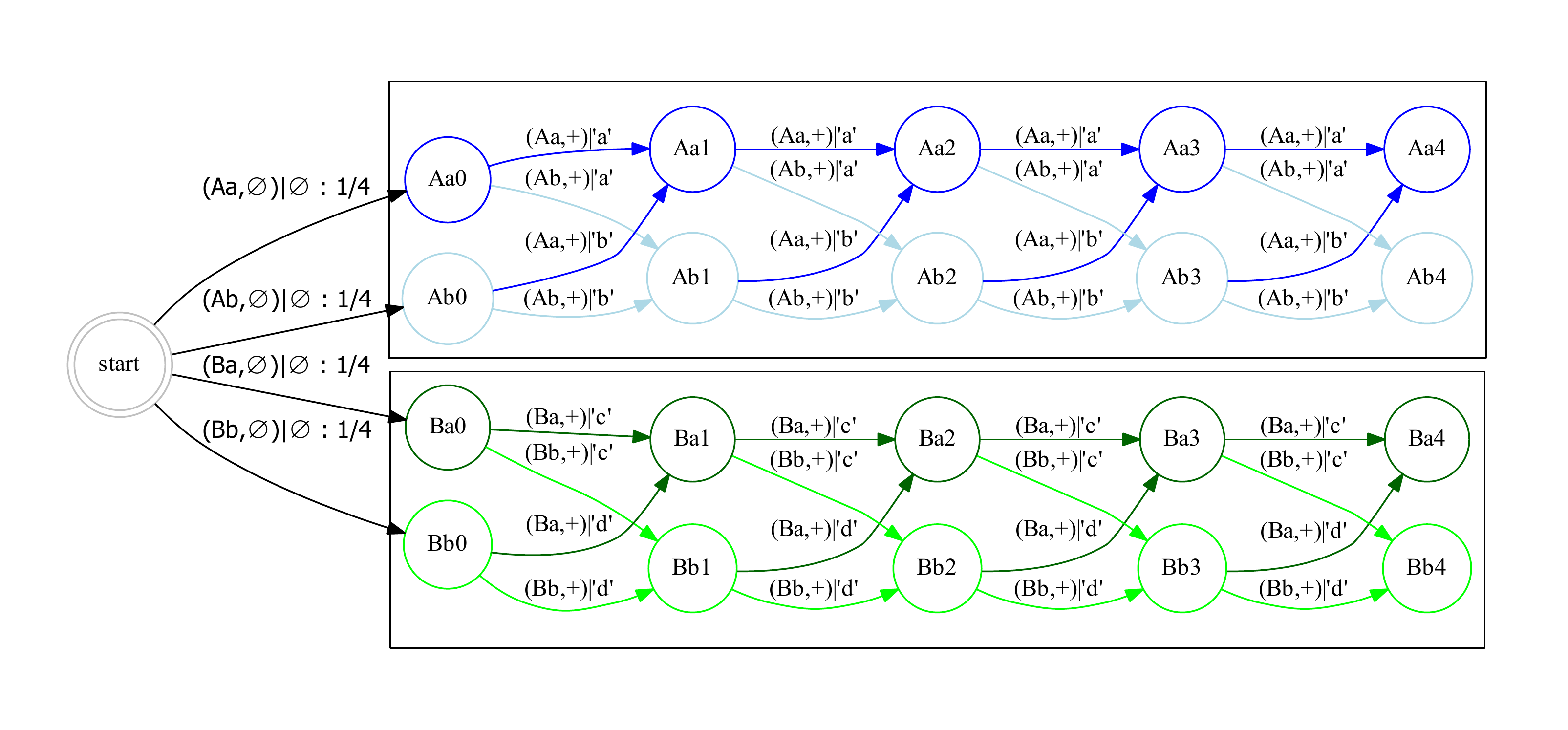}
  \caption{}
  \label{sfig:eps2}
\end{subfigure}
\captionsetup{singlelinecheck=off}
\caption[]{A simplified visualization of $\epsilon$-transducers for two consecutive tasks A.2.1 (\subref{sfig:eps1}) and A.2.2 (\subref{sfig:eps2}). No error or instance switch transitions are shown. One can see the apparent possible sub-structures that the two tasks share, that could be exploited by a gradual learner. The \textit{state} naming convention is as follows:
\begin{itemize}
\item First (capital) letter denotes an instance
\item The number at the end of the state descriptor represents the number of consecutive positive rewards the agent has received
\item The middle letter (only in the Subfigure~\subref{sfig:eps2}) defines from which group a letter was provided by the environment for the next step
\end{itemize}
For these visualizations, five consecutive positive rewards are required to complete an instance (fully shown in Figure~\ref{fig:eps4}).}
\label{fig:simple}
\end{figure}

\clearpage

\begin{figure}
\centering
\includegraphics[clip, trim=0cm 0cm 0cm 0cm, height=0.95\textheight]{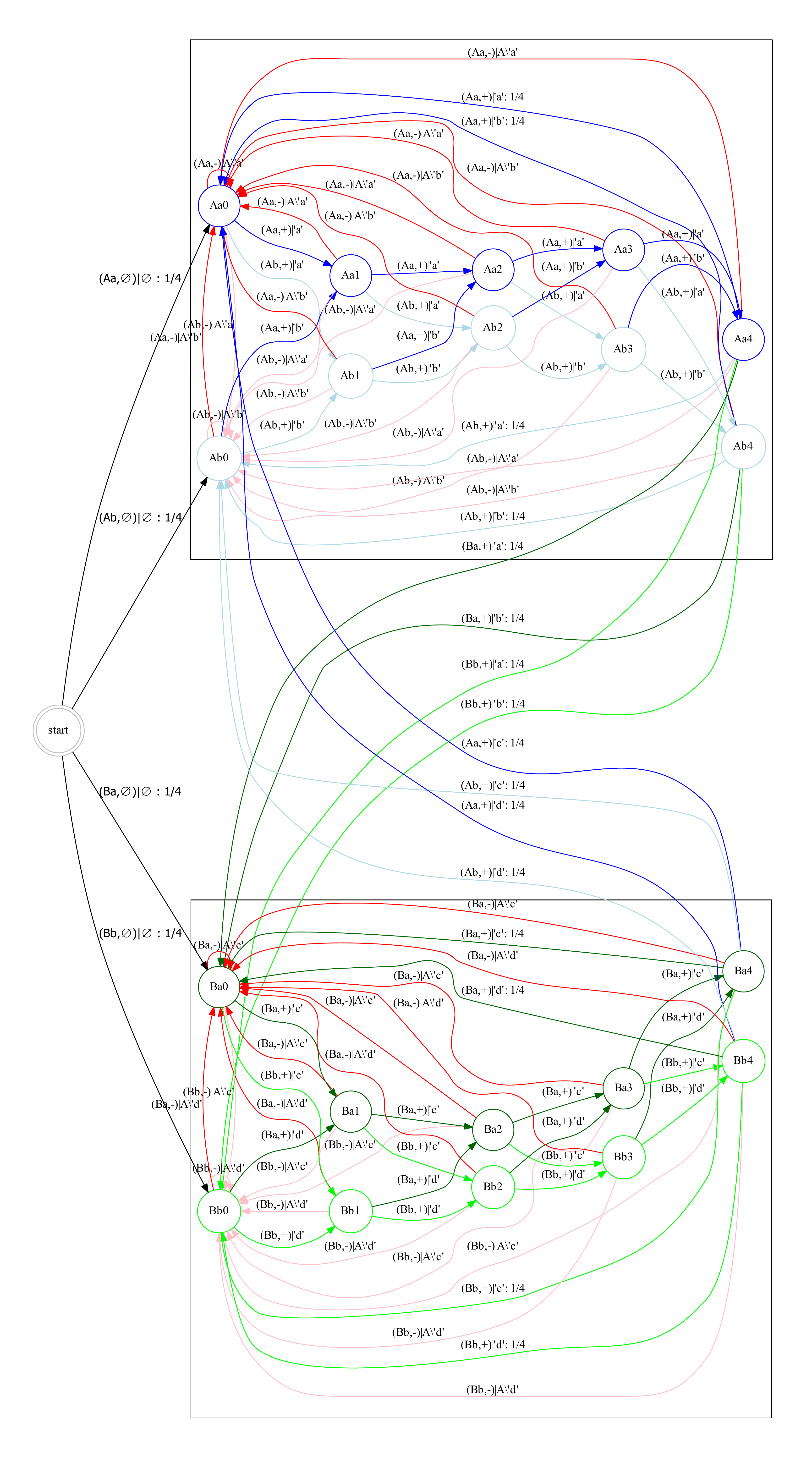}
\caption{Full visualization of $\epsilon$-transducer for task A.2.2 from the challenge curriculum. Same state naming convention applies as in Figure~\ref{fig:simple}.}
\label{fig:eps4}
\end{figure}

\end{document}